\documentclass{article}

\usepackage{aaai}
\usepackage{epsfig}

\newcommand{\commadots}[0]{,\ldots ,}
\def\AND     { \,\wedge\,		  }
\def\OR      { \,\vee\,			  }

\newcommand{\PSPACE}{{\rm PSPACE}} 
\newcommand{\SAT}[0]{\mbox{\sc sat}}
\newcommand{\QUIP}[0]{\mbox{\sf QUIP}}
\newcommand{\QSAT}[1]{$\mbox{\sc qsat}_{#1}$}
\newcommand{\NM}[0]{\mbox{\sc nmp}}
\newcommand{\iec}[0]{i.e.,\ }
\newcommand{\egc}[0]{e.g.,\ }

\newcommand{\nop}[1]{}

\newcommand{\limplies}{\rightarrow}
\newcommand{\lequiv}{\leftrightarrow}

\newcommand{\NP}{{\rm NP}}

\newcommand{\CONP}{\mbox{\rm co-}\NP }

\newcommand{\nm}{nonmonotonic}

\newcommand{\quip}{\texttt{quip}}

\newcommand{\nega}{\texttt{!}}
\newcommand{\cona}{\texttt{\&}}
\newcommand{\cons}{\texttt{|=}}

\begin{document}

\title{\QUIP---A Tool for Computing Nonmonotonic Reasoning 
Tasks\thanks{This work was partially supported by 
the Austrian Science Fund Project N Z29-INF.}
}

\author{Uwe Egly, Thomas Eiter,  Hans Tompits, and  Stefan Woltran \\
Technische Universit\"at Wien \\
	Abt.\ Wissensbasierte Systeme 184/3 \\
	Favoritenstra{\ss}e~9--11,
	A--1040 Wien, Austria \\
	e-mail:  [uwe,eiter,tompits,stefan]@kr.tuwien.ac.at
}
            
\maketitle

\begin{abstract}
\noindent
In this paper, we outline the prototype of an automated inference tool, 
called \QUIP,  which provides a uniform implementation for several 
nonmonotonic reasoning formalisms. The theoretical basis of \QUIP\ is 
derived from well-known results about the computational complexity of 
nonmonotonic logics and exploits a representation of the different 
reasoning tasks in terms of {\em quantified boolean formulae\/} (QBFs). 
\end{abstract}

\section{General Information}\label{sec:intro}


Designing theorem provers for nonmonotonic reasoning formalisms is challenged 
by the increased computational complexity compared to the inherent complexity 
of classical reasoning.
As is well-known~\cite{Gottlob:1992,Eiter:Gottlob:1993,eite-gott-95,eite-gott-92bx}, 
the main reasoning tasks for almost all propositional nonmonotonic logics are either $\Sigma_2^p$-complete or $\Pi_2^p$-complete, \iec they are located at the second 
level of the polynomial hierarchy, whereas classical reasoning resides at the first
level of that hierarchy.

Important constituents in establishing the $\Sigma_2^p$-completeness results
for the different nonmonotonic reasoning tasks are so-called {\em quantified 
boolean formulae\/} (QBFs), which generalize ordinary propositional formulae by 
the admission of quantifiers ranging over propositional variables.
To wit, demonstrating $\Sigma_2^p$-hardness of a given nonmonotonic decision 
problem, say \NM,  is usually achieved by reducing  the problem \QSAT{2} to \NM, 
where \QSAT{2} is the problem of deciding the truth of QBFs being in prenex 
normal-form whose leading quantifiers are ordered like $\exists P\,\forall Q$ ($P,Q$ 
are lists of propositional variables), which  is complete for $\Sigma_2^p$. 
(In general, for any class $\Sigma_i^p$, $i\geq 1$, there is a corresponding 
decision problem, \QSAT{i}, complete for $\Sigma_i^p$, whose task is to check the 
truth of QBFs of the form 
$\exists P_1\,\forall P_2\ldots \exists P_{n-1}\,\forall P_n\; \Phi$, 
where $P_1\commadots P_n$ are lists of propositional variables and $\Phi$ 
is a propositional formula.) More precisely,  $\Sigma_2^p$-hardness of \NM\ is shown by
constructing a (polynomial) transformation $\cal S$ such that for every instance $I$ 
of \QSAT{2} it holds that $I$ is a yes-instance of \QSAT{2} iff ${\cal S}(I)$ is a 
yes-instance of \NM.

Now, given the membership of a nonmonotonic reasoning task \NM\ in the class 
$\Sigma_2^p$, and from the $\Sigma_2^p$-hardness of \QSAT{2}, it follows 
immediately that every such decision problem \NM\ can be reduced to \QSAT{2}, 
\iec there is a (polynomial) transformation $\cal T$ such that for every instance 
$I$ of \NM\ it holds that $I$ is a yes-instance of \NM\ iff ${\cal T}(I)$ is a 
yes-instance of \QSAT{2}.

In this paper, we describe the prototype system
\QUIP~\cite{egly-eiter-etal:wlp00,egly-eiter-etal:aaai00}, an automated 
reasoning tool utilizing transformations of the latter kind to implement 
several different reasoning formalisms. The basic idea is to employ 
\emph{existing sophisticated theorem provers for quantified boolean formulae}, 
taking care of evaluating the resultant instances of \QSAT{2}.

At present, \QUIP\ handles the following propositional nonmonotonic reasoning
approaches:

\begin{itemize}
\item abduction;
\item autoepistemic logic;
\item default logic;
\item disjunctive stable model semantics;
\item circumscription.
\end{itemize}
The system has been implemented in C using standard tools like LEX and YACC 
(comprising a total of 2000 lines of code, excluding the used QBF-solver); 
it runs currently in a Unix environment (Sun/Solaris and Linux), but is 
easily portable to other operating systems as well. 

The use of QBFs expressing advanced reasoning tasks has been advocated in~\cite{cado-etal-98,rint-99} and is a natural generalization of a 
similar method applied for problems in NP. Such problems are often solved 
by a reduction to \SAT, the satisfiability problem of classical propositional 
logic, which is \NP-complete (see, \egc 
\cite{aaai96-2*1194,giunchiglia-sebastiani:1996a}). Besides the applications 
discussed in this paper, a reduction of planning problems to QBFs has been 
given in \cite{rintanen:1999:plans}. 

In general, since the evaluation of \emph{arbitrary} QBFs is \PSPACE-complete 
(in contrast to the $\Sigma_i^p$-completeness of the restricted QBFs mentioned 
above), in principle \emph{any} formalism having a decision problem in \PSPACE\ 
can be handled by \QUIP, provided a proper transformation has been found.

The next section outlines the overall architecture of \QUIP\ and gives some 
background information on the different reasoning tasks implemented in \QUIP. 
Then, the usability of the approach is described. The last section contains 
a discussion on benchmark problems and a comparison with other implementations.

\section{Description of the System}\label{sec:description} 

\subsection{System Architecture}
\begin{figure*}
\begin{center}
\psfig{file=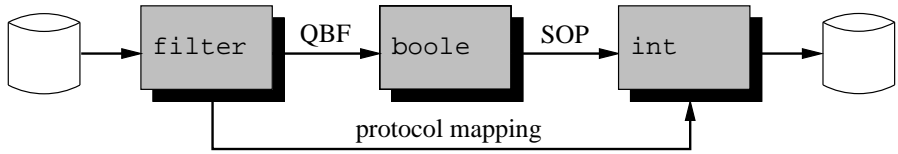}
\end{center}
\caption{\QUIP's system architecture.\label{fig:quip-arch}}
\end{figure*}

The overall architecture of \QUIP\ is depicted in Figure~\ref{fig:quip-arch}. 
\QUIP\ consists of three parts, namely the {\tt filter} program, the 
QBF-evaluator {\tt boole}, and the output interpreter {\tt int}. 

The input filter translates the given problem description (\egc a default 
theory, a disjunctive logic program, an abductive problem, etc.) into 
a quantified boolean formula, which is then sent to the QBF-evaluator 
{\tt boole}. The result of {\tt boole}, usually a formula in disjunctive 
normal form (often called \emph{sum of products}, SOP), is interpreted 
by {\tt int}. The latter part associates a meaningful interpretation 
to the formulae occurring in SOP and provides an explanation in terms 
of the underlying problem instance (\egc an extension, a stable model, 
abductive explanations, etc.). The interpretation relies on a mapping 
of internal variables of the generated QBF into concepts of the problem 
description which is provided by {\tt filter}.

The QBF-evaluator {\tt boole} is a commonly available propositional 
theorem prover based on \emph{binary decision diagrams} 
(BDDs)~\cite{Bryant:1986}.\footnote{For a comparison between 
different BDD packages, cf.~\cite{Long:1998}.} 
We chose to use this particular tool because of several reasons. For one, 
the program, together with its source code, is in the public domain and 
can be downloaded from the web 
(see  http://www.cs.cmu.edu/$\sim$modelcheck/bdd.html).
Second, it represents a sophisticated reasoning engine with several 
years of development. Third, it does not require of the input 
formula to be in a specific normal form. The latter point distinguishes 
{\tt boole} from other QBF-solvers,  like those proposed 
in~\cite{cado-etal-98,rint-99}, which operate on formulae being in 
\emph{prenex conjunctive normal form}. Although these provers can 
in principle also be employed in \QUIP, their inclusion would require 
an additional normal form translation, since the ``natural'' reductions 
of nonmonotonic reasoning tasks to QBFs do in general \emph{not} result 
in formulae being in a particular normal form.

In order to incorporate new formalisms into \QUIP, one has to extend the 
{\tt filter} program responsible for the appropriate reductions, the 
mapping of the variables, and the interpreter {\tt int}. The deductive 
engine remains unchanged in this process.

\subsection{Implemented Reasoning Tasks}

In this subsection, we briefly discuss the different reasoning tasks 
currently implemented in \QUIP. We will not present the concrete 
transformations into QBFs expressing these tasks; details are given
in~\cite{egly-eiter-etal:aaai00}.

All formalisms processed by \QUIP\ are propositional. In what follows, 
we assume a propositional language $\cal L$ generated by a finite set of
propositional variables $V$ using the standard sentential connectives 
$\neg$, $\AND$, $\OR$, $\limplies$, and $\lequiv$.
A {\em theory} is a finite set $T \subseteq {\cal L}$, which will be 
identified with the formula $\bigwedge_{\phi \in T} \phi$.

Note that, strictly speaking, the different decision problems presented 
below return either \emph{yes} or \emph{no}, but usually one is more 
interested in the corresponding \emph{function problem} returning the 
actual objects responsible for a \emph{yes} answer (like the extensions 
of a default theory, the stable models of a logic program, etc.). \QUIP\ 
permits queries of this form, by engaging the so-called \emph{detail mode} 
(see discussion below).

\paragraph{Abduction.} 

Classical abduction from a theory $T$ on $V$ may be defined as
follows~\cite{selm-leve-90,pool-89,eite-gott-92bx}. 
Let $H \subseteq V$ be a set of {\em hypotheses}, and let $p \in V$ be 
a distinguished atom. A subset $E \subseteq H$ is an {\em abductive 
explanation} for $p$ from $T$ and $H$, if 
\begin{description}
 \item[\hphantom{i}{\rm (i)}] $T \cup E$ is consistent, and 
 \item[{\rm (ii)}] $T\cup E \models p$, \iec $T\cup E$ logically implies $p$.
\end{description}
An explanation $E$ is {\em minimal}, if no proper subset $E' \subset
E$ is an abductive explanation. 

The following tasks are implemented in \QUIP:

\begin{itemize}
\item Given a theory $T$, a set $H$ of variables, and an atom $p$. 
Is there a (minimal) abductive explanation $E\subseteq H$ for $p$ 
from $T$ and $H$?

\item The \emph{relevance problem}: Given a theory $T$, a set $H$ of 
variables, an atom $p$, and some hypothesis $h\in H$. Is there a 
(minimal) abductive explanation $E\subseteq H$ for $p$ from $T$ and 
$H$ containing $h$?

\item The \emph{necessity problem}: Given a theory $T$, a set $H$ 
of variables, an atom $p$, and some hypothesis $h\in H$. Does $h$ 
occur in every (minimal) abductive explanation $E\subseteq H$ for 
$p$ from $T$ and $H$?

\end{itemize}

\paragraph{Autoepistemic logic.}
The language of Moore's autoepistemic logic \cite{moor-85} contains the
modal operator $L$, where $L\phi$ intuitively means that $\phi$ is
believed. By ${\cal L}_L$ we denote the language ${\cal L}$ extended
by $L$.  Formulae $L\phi$ are viewed as propositional
variables, which are called {\em modal atoms}.

A \emph{stable expansion} of an autoepistemic theory $T \subseteq 
{\cal L}_L$ is a set  $E\subseteq {\cal L}_L$ such that
$$E = Th(T \cup \{ L\phi \mid \phi \in E) \cup \{ \neg L\phi \mid
\phi \notin E\}),$$
where $Th(\cdot)$ is the classical consequence operator with respect 
to the extended language ${\cal L}_L$. 

\QUIP\ handles the following tasks:

\begin{itemize}
\item Given an autoepistemic theory $T \subseteq {\cal L}_L$, is
 there a stable expansion $E$ of $T$?

\item \emph{Brave reasoning}: Given an autoepistemic theory 
$T \subseteq {\cal L}_L$ and some formula $\Phi$, is there a 
stable expansion $E$ of $T$ containing $\Phi$?

\item \emph{Skeptical reasoning}: Given an autoepistemic theory 
$T \subseteq {\cal L}_L$ and some formula $\Phi$, is $\Phi$ 
contained in every stable expansion $E$ of $T$?
\end{itemize}
For brave and skeptical reasoning, if the detail mode is 
engaged, \QUIP\ returns \emph{witnesses} corresponding to 
theses tasks: for brave reasoning, all stable expansions
containing $\Phi$ are returned, whereas for skeptical reasoning, 
all stable expansions \emph{not} containing $\Phi$ are returned.

\paragraph{Default logic.}

A \emph{default theory} is a pair $\Delta = (T,D)$, where 
$T\subseteq {\cal L}$ is a set of formulae and $D$ is a set 
of \emph{defaults} of the form $\frac{\alpha~:~\beta}{\gamma}$.%
\footnote{For simplicity, we omit multiple justifications here.}  
Intuitively, the default is applied ($\gamma$ is concluded) if 
$\alpha$ is provable and the {\em justification} $\beta$ can be consistently
assumed.

The semantics of $\Delta=(T,D)$ is defined in terms of
\emph{extensions}~\cite{reit-80}. Following~\cite{mare-trus-93a}, 
extensions can be characterized thus.
For any $S \subseteq {\cal L}$, let $D(S)$ be the monotonic rules 
$\{ \frac{\alpha}{\gamma} \mid \frac{\alpha~:~\beta}{\gamma} \in D, 
\neg \beta \notin S\}$. Then, $E\subseteq {\cal L}$ is an extension 
of $\Delta$ iff $ E = Th^{D(E)}(T)$, where $Th^{D(E)}(T)$ is the set 
of all formulae derivable from $T$ using classical logic together 
with the rules from $D(E)$.

\QUIP\ expresses the following reasoning tasks:

\begin{itemize}
\item  Given a default theory $\Delta$, is there an extension 
$E$ of~$\Delta$?

\item \emph{Brave reasoning}: Given a default theory $\Delta$ and 
some formula $\Phi$, is there an extension $E$ of $\Delta$ 
containing~$\Phi$?

\item \emph{Skeptical reasoning}: Given a default theory $\Delta$ and 
some formula $\Phi$, is $\Phi$ contained in every extension $E$ of 
$\Delta$?

\end{itemize}

As for stable expansions, \QUIP\ returns witnesses if the detail 
mode is engaged. Moreover, each of these reasoning tasks has been 
implemented in terms of two independent transformations: The first 
category of reductions is based on the characterization of extensions 
discussed above; the second category is based on a characterization of 
extensions using the notion of a \emph{full set}~\cite{niem-95}. 
Interestingly, although the transformations based on the latter 
method are more succinct than the corresponding reductions of the 
first kind, they have often an inferior performance compared to 
the former transformations.

\paragraph{Disjunctive Logic Programming.}

A \emph{disjunctive logic program}, $\Pi$, is a set of rules 
$$r:\quad H(r) \leftarrow P(r),N(r)$$ where $H(r)$ is a disjunction of
variables, $P(r)$ is a conjunction of variables, and $N(r)$ is a
conjunction of negated variables.  A Herbrand interpretation $I$ of
$V$ is a {\em stable model} of $\Pi$ \cite{gelf-lifs-88,przy-91}, if it
is a  minimal model (with respect to set-inclusion) of the program $\Pi^I$
resulting from $\Pi$ as follows: remove each rule $r$ such that $I
\models a$
for some $\neg a$ in $N(r)$, and remove $N(r)$ from all
remaining clauses.

Similar to autoepistemic logic and default logic, \QUIP\ handles the 
problem whether a given logic program has a stable model, as well as 
brave and skeptical reasoning.

\paragraph{Circumscription.} In contrast to the formalisms described 
above, propositional circumscription is already a quantified boolean 
formula, hence it  does not require a separate reduction. So, \QUIP\ 
can handle circumscription in a straightforward way.

In the propositional case, the parallel circumscription of a set of 
atoms $P=\{p_1,\ldots,p_n\}$ in a theory $T$, where the atoms
$Q$ are fixed and the remaining atoms 
$Z=\{z_1,\ldots,z_m\}=V\setminus (P\cup Q)$ may vary,
is given by the following QBF $\mathit{CIRC}(T;P,Z)$, cf. 
\cite{lifs-85a}:
$$
T \land \forall P'\, \forall Z'\Big((T[P/P',Z/Z'] 
\land (P' \leq P)) \limplies (P\leq P')\Big).
$$
Here, $P'=\{p'_1,\ldots,p'_n\}$ and $Z'=\{z'_1,\ldots,z'_m\}$ are sets
of new propositional variables corresponding to $P$ and $Z$,
respectively, and $T[P/P',Z/Z']$ results from $T$ by substitution of the
variables in $P'\cup Z'$ for those in $P\cup Z$. \QUIP\ implements 
circumscriptive inference of a formula $\phi$  from $T$, which is 
expressed by the QBF
$$\forall V(\mathit{CIRC}(T;P,Z) \limplies \phi).$$

\section{Applying the System}\label{sec:application}


\subsection{Methodology}

                                        

One of the basic motivations for the development of \QUIP\ was to 
make a rapid prototyping tool available, aimed for experimenting 
with different knowledge-represen\-tation formalisms.
Accordingly, \QUIP\ is designed in such a way that the important 
reasoning tasks corresponding to the different formalisms under 
consideration can be directly encoded as queries.
So, the methodology of specifying queries suitable to be processed 
by \QUIP\ is  the same as the methodology of formalizing a particular 
problem with one of the formalisms implemented in \QUIP\ (subject to 
the restriction, of course, that \QUIP\ currently accepts only queries 
specified over a propositional language; but in a future version it is 
planned to extend the language to allow function-free formulae with 
variables as well.) 

\begin{table*}
\begin{center}
\begin{tabular}{l||r|r|r|r|r|r|r|r|r|r|r|r|r|r}
$k$               &    2 &     3 &     4 &     5 &     6 &     7 &     8 &    9 &   10 &   11 &   12 &  13 &    14 &  15
\\\hline
dlv                &  0.0 &   0.0 &   0.0 &   0.1 &   0.2 &   0.4 &   0.7 &  1.1 &  2.0 &  3.3 &  5.2 &  8.3 & 12.5 &  19.3
\\
\QUIP\ (DLP)       &  0.4 &   0.4 &   0.4 &   0.4 &   0.6 &   0.4 &   0.5 &  0.6 &  0.9 &  1.8 &  3.8 &  9.0 & 23.7 & 54.1
\\\hline
DeReS              &  0.0 &   0.2 &   1.7 &  21.2 &  67.9 &    -- &    -- &   -- &   -- &   -- &   -- &   -- &  --  & --
\\
\QUIP\ (DL)        &  0.0 &   0.0 &   0.1 &   0.1 &   0.1 &   0.2 &   0.3 &  0.5 &  1.0 &  2.1 &  4.8 & 12.0 & 33.5 & 80.4
\\\hline
\QUIP\ (ABD)       &  0.0 &   0.0 &   0.0 &   0.0 &   0.0 &   0.0 &   0.1 &  0.2 &  0.5 &  1.5 &  4.2 & 13.6 & 48.7 & --
\end{tabular}
\end{center}
\caption{Performance Results
\label{table:dl1}}
\end{table*}

\subsection{Specifics}\label{sec:methodology}
                                        

Let us illustrate how the queries of \QUIP\ are structured. Generally, 
\QUIP\ takes an input file as argument and writes the output
to standard-out: 
{\small\begin{quote}
\texttt{\quip\ }\em{input\_file} 
\end{quote}}
\noindent
The input file comprises \nm\ theories and specifications of 
reasoning tasks. The format of the input uses two major concepts: 
\emph{definitions} and \emph{commands}. In the definitions, one 
can specify abductive theories, autoepistemic theories, default 
theories, logic programs, and propositional theories in general. 
The commands specify which reasoning tasks (with respect to the 
chosen formalism) have to be executed.
Theories and formulas can be nested by using suitable names, so 
definitions can be edited rather conveniently.
Further, commands can refer to specified theories without the 
need to describe them more than once.
An input file can contain several definitions and commands, 
even referring to different formalisms.

In the following we describe some of these features using a 
simple example from default logic. Consider the following 
default theory $\Delta=(T,D)$, representing the well-known 
Nixon-diamond:
\begin{eqnarray*}
T & = & \{ \emph{Republican}, \emph{Quaker} \}; \\
D & = & \left\{ \frac{\emph{Republican} : \neg 
        \emph{Pacifist}}{\neg \emph{Pacifist}}, \frac{\emph{Quaker} :
        \emph{Pacifist}}{\emph{Pacifist}} \right\}. 
\end{eqnarray*}
This default theory has two extensions: 
\begin{quote}
$Th( \{\emph{Republican}, \emph{Quaker}\} \cup 
\{\neg \emph{Pacifist}\})$;

$Th( \{\emph{Republican}, \emph{Quaker}\} \cup 
\{\emph{Pacifist}\})$. 
\end{quote}

The extensions of this example can be computed by \QUIP\ 
using the following input file, named \texttt{nixon}:
{\small\begin{verbatim}
   @SET DETAIL
      T :=   Republican & Quaker
   @D D := { Republican : !Pacifist | !Pacifist, 
             Quaker : Pacifist | Pacifist 
           } 

   @DL ( T ; @D D )
\end{verbatim}}
\noindent
Executing the command
{\small\begin{quote}
\texttt{\quip\ nixon} 
\end{quote}}\noindent
results in the following output:
{\small\begin{quote}
\texttt{Th( \{(Republican)\cona(Quaker)\} u \{(Pacifist)\} ) } \\
\texttt{Th( \{(Republican)\cona(Quaker)\} u \{(!Pacifist)\} ) }
\end{quote}}

The meaning of the different commands and definitions in the 
input file \texttt{nixon} can be explained as follows. First 
of all, the command \texttt{@SET DETAIL} invokes the detail 
mode, \iec all extensions will be displayed. (Recall that \QUIP\ 
admits the processing of both \emph{decision problems} and 
\emph{function problems}.) Choosing the command \texttt{@UNSET
DETAIL} would have resulted in a simple yes/no answer.

The next tokens of the input file specify the constituents of 
the default theory~$\Delta$: \texttt{T} represents the background 
knowledge of $\Delta$, and \texttt{D} contains the defaults. To 
avoid ambiguities, references to defaults always have to start 
with a special string ``\texttt{@D}''. The logical operators are
represented in the obvious way (``\cona'' denotes conjunction, 
``\nega'' negation, and ``\texttt{|}'' separates a default's 
consequent from its justification). 

The command \texttt{@DL} tells \QUIP\ to compute the extensions 
of the specified default theory. To perform brave or skeptical 
reasoning, the input file \texttt{nixon} would be changed as follows:
{\small\begin{verbatim}
   @SET DETAIL
      T :=   Republican & Quaker
   @D D := { Republican : !Pacifist | !Pacifist, 
             Quaker : Pacifist | Pacifist 
           } 

   @SET BRAVE
   @DL ( W ; @D D ) |= Pacifist

   @SET SKEPTICAL
   @DL ( W ; @D D ) |= Pacifist

\end{verbatim}} 
The \texttt{@SET}-commands switches between the two reasoning modes; 
``\cons'' is the symbol standing for the respective default 
consequence relations. The first command checks whether 
\emph{Pacifist} is a brave consequence of $\Delta$, the 
second commands checks skeptical consequence of \emph{Pacifist} 
from $\Delta$. Observe that it is permitted that an input file 
contains a sequence of reasoning tasks.

The output corresponding to the first command will be
{\small\begin{quote}
\texttt{Th( \{(Republican)\cona(Quaker)\} u \{(Pacifist)\} ) }
\end{quote}}\noindent
while the second command displays the extensions from 
which \emph{Pacifist} does \emph{not} follow: 
{\small\begin{quote}
\texttt{Th( \{(Republican)\cona(Quaker)\} u \{(!Pacifist)\} ) }
\end{quote}}

\subsection{Users and Usability} 

Obviously, proper usage of \QUIP\ depends on a potential user's 
ability to express a given problem in terms of the implemented 
formalisms. However, a particular advantage of \QUIP\ is that it 
incorporates several different knowledge representation formalisms. 
Hence, users have a \emph{choice} selecting among different 
(yet closely related) approaches, singling out that particular 
formalism best suited for a specific purpose, or choosing a method 
based on a user's personal preference (\egc because he/she understands 
that particular approach best). As well, the problem can be 
represented with different methods \emph{simultaneously}, specifying 
the instances of the resultant formalizations in a single input file, 
which can then be processed by \QUIP\ requiring only one initial 
execution command.

\section{Evaluating the System}\label{sec:eval}


\subsection{Benchmarks}

There are different approaches how systems handling knowledge 
representation formalisms can be evaluated. One is to perform 
comparisons taking into account the representational power of 
the implemented formalisms. That is to say, under such a 
comparison, one chooses some ``natural'' problems, encodes it 
with respect to the specific methodologies associated with the 
implemented formalisms, and uses the resultant instances as 
queries of the respective systems. So, basically, different 
systems are compared on the basis of (possibly) different 
representations \emph{of the same problem}. However, to achieve 
a fair comparison, it is necessary that the experimenter is able 
to encode the given problem ``in the best possible way'' with 
respect to the particular formalisms. Also, incompatibilities 
of the underlying formalisms render comparisons between different 
systems often difficult.

Another possibility is to compare systems on  problem classes 
\emph{common} to each of the considered systems. Such a 
comparison is appropriate if different implementations of 
\emph{the same formalism} are evaluated.

Since \QUIP\ incorporates a wide array of different knowledge 
representation methods, comparisons with other implementations 
can in general be achieved employing the latter procedure.

Accepted benchmarks for nonmonotonic theorem provers have 
been realized by the well-known TheoryBase system~\cite{theorybase}. 
This test-bed provides encodings of various graph problems in terms 
of default theories or equivalent logic programs. However, the 
generated problems are at most \NP-hard (or \CONP-hard, depending 
on the reasoning task), and thus do not take full advantage of the 
expressibility supported by the nonmonotonic formalisms. The 
only practically applied benchmark test utilizing 
a $\Sigma_2^p$-hard problem is the strategic companies example 
carried out for testing the system \texttt{dlv}~\cite{eite-etal-98a}.  

Here, we propose a straightforward method how $\Sigma_2^p$-hard 
benchmark problems for propositional nonmonotonic reasoning 
formalisms can be generated. The idea is to use the class of 
problems establishing the $\Sigma_2^p$-hardness of a formalism. 
Recall from our previous discussion that  $\Sigma_2^p$-hardness 
of a nonmonotonic formalism is usually demonstrated by constructing 
a (polynomial) transformation $\cal S$ mapping instances $I$ of 
\QSAT{2} (the evaluation problem of QBFs having quantifier order 
$\exists\,\forall$) into instances ${\cal S}(I)$ of the considered 
nonmonotonic reasoning task, \NM, such that $I$ is a yes-instance 
of \QSAT{2} iff ${\cal S}(I)$ is a yes-instance of \NM. Thus, in 
some sense, the class of problems ${\cal S}(I)$ represents 
``worst-case'' examples for the problem \NM, and therefore is 
particularly useful estimating the performance of a theorem prover 
solving the task \NM. Moreover, these problems are easily scalable 
by parameterizing different instances of \QSAT{2}. An added feature 
of \QUIP\ is that these examples provide at the same time a simple 
method for \emph{testing} whether the implementation works correct, 
because \QUIP\ turns instances ${\cal S}(I)$ of \NM\ back to 
instances ${\cal T}({\cal S}(I))$ of \QSAT{2}, satisfying the 
condition that $I$ is a yes-instance of \QSAT{2} iff  
${\cal T}({\cal S}(I))$ is a yes-instance of \QSAT{2}. 
The next subsection describes comparisons between \QUIP\ 
and some state-of-the-art provers on the basis of these 
benchmark problems.

\subsection{Comparison} 


We compare the default-logic module of \QUIP\ with 
DeReS~\cite{chol-etal-96} and the logic-programming module 
of \QUIP\ with \texttt{dlv}~\cite{eite-etal-98a}, using the class 
of examples discussed above. Space limits preclude a discussion on 
the structure of these problems; details can be found in the relevant 
literature (\egc \cite{Gottlob:1992,Eiter:Gottlob:1993,eite-gott-95,eite-gott-92bx}). 
We do not include a comparison with \texttt{smodels}~\cite{niem-simo-96} 
here, because that system is currently not designed to handle 
$\Sigma_2^p$-problems (a comparison between \QUIP, DeReS, \texttt{dlv}, 
\texttt{smodels}, and Theorist~\cite{pool-89}, using examples from 
TheoryBase and some abduction problems, is given in 
\cite{egly-eiter-etal:aaai00}).

Results of the comparison are given Table~\ref{table:dl1}. All tests 
have been performed on a SUN ULTRA 60 with 256MB RAM; the run-time 
is measured in seconds with an upper limit of 90sec (\iec instances
requiring a longer period are not displayed). The first group of entries 
gives the results for the disjunctive logic programming test; the second 
group gives the results for the default logic test; and the final row 
contains some measurements using instances of the corresponding abductive 
problem class. The respective input-QBFs have been randomly generated 
and are parameterized by the number $k$ of existential quantifiers (the 
number of variables was held fixed and was set to 20). Although the 
given results represent only a small sample, they do indicate that 
our \emph{ad hoc} implementation performs sufficiently well. 

\subsection{Problem Size} 


It is rather obvious that \QUIP\ cannot compete with 
state-of-the-art implementations like \texttt{dlv} or 
\texttt{smodels} in terms of problem size. These tools 
are highly optimized systems developed with a particular 
semantics in mind, whereas the purpose of \QUIP\ is to 
provide a \emph{uniform} method dealing with several 
knowledge representation tasks at the same time. Under 
this perspective, and taking into account that \QUIP\ 
utilizes at present no optimizations whatsoever, our 
results demonstrate that implementing nonmonotonic reasoning
formalisms using reductions to  quantified boolean formulae 
is a feasible approach. Moreover,  the modular architecture 
of \QUIP\ allows an easy scalability and parallelization, 
by using, \egc several QBF-provers  simultaneously, each of 
which with its own optimization method.

\bibliographystyle{aaai}

\end{document}